\documentclass[10pt]{article}
\usepackage{fancyhdr}
\usepackage{extramarks}
\usepackage{amsmath}
\usepackage{amsthm}
\usepackage{amsmath}
\usepackage{amsfonts}
\usepackage{siunitx}
\usepackage{tikz}
\usepackage{float}

\usepackage{algorithm}
\usepackage{algpseudocode}
\usepackage[all]{xy}
\usepackage{tikz-cd}
\usepackage{multirow}
\usepackage{booktabs}
\usepackage{graphicx,subfigure}
\usepackage{palatino}
\usepackage{graphicx}
\usepackage{subfigure}
\usepackage[colorlinks,linkcolor=black,anchorcolor=black,citecolor=black,urlcolor=blue]{hyperref}
\usepackage{amsmath,bm}
\DeclareSymbolFont{toneitalic}{T1}{\familydefault}{m}{it}
\DeclareMathSymbol{\cpartial}{\mathord}{toneitalic}{"F0}
\usepackage{booktabs}
\usepackage{mathtools}
\usepackage{amssymb}
\usepackage{caption}
\usepackage{capt-of}
\usepackage{mciteplus}
\usepackage{cite}
\usepackage{mathrsfs}
\usepackage{parskip}
\usepackage{soul}
\usepackage{textcomp}
\usepackage[colaction]{multicol}
\usepackage[switch]{lineno}
\usepackage{lipsum}
\usepackage{etoolbox}
\usepackage{longtable}
\usepackage{array}
\usepackage{tablefootnote}
\usepackage{ragged2e}
\usepackage{lscape}
\newcolumntype{C}[1]{>{\centering\arraybackslash}p{#1}}
\usetikzlibrary{automata,positioning}
\usetikzlibrary{automata,positioning}

\usepackage{tikz,xcolor,hyperref}
\definecolor{lime}{HTML}{A6CE39}
\DeclareRobustCommand{\orcidicon}{%
	\begin{tikzpicture}
	\draw[lime, fill=lime] (0,0)
	circle [radius=0.16]
	node[white] {{\fontfamily{qag}\selectfont \tiny ID}};
	\draw[white, fill=white] (-0.0625,0.095)
	circle [radius=0.007];
	\end{tikzpicture}
	\hspace{-2mm}
}
\foreach \x in {A, ..., Z}{%
	\expandafter\xdef\csname orcid\x\endcsname{\noexpand\href{https://orcid.org/\csname orcidauthor\x\endcsname}{\noexpand\orcidicon}}
}

\begin{document}
\setlength\parindent{0pt}
\title{Persistent Laplacian-enhanced Algorithm for Scarcely Labeled  Data Classification}

\author{Gokul Bhusal$^1$,
Ekaterina Merkurjev$^{1,2}$\footnote{Corresponding author,
	Email:  merkurje@msu.edu} ~ and
Guo-Wei Wei$^{1,3,4}$\\
$^1$ Department of Mathematics, \\
Michigan State University, MI 48824, USA.\\
$^2$ Department of Computational Mathematics, Science and Engineering\\
Michigan State University, MI 48824, USA.\\
$^3$ Department of Electrical and Computer Engineering,\\
Michigan State University, MI 48824, USA. \\
$^4$ Department of Biochemistry and Molecular Biology,\\
Michigan State University, MI 48824, USA. \\
}
\date{}
\maketitle
\abstract{

The success of many machine learning (ML) methods depends crucially on having large amounts of labeled data. However, obtaining enough labeled data can be expensive, time-consuming, and subject to ethical constraints for many applications. One approach that has shown tremendous value in addressing this challenge is semi-supervised learning (SSL); this technique utilizes both labeled and unlabeled data during training, often with much less labeled data than unlabeled data, which is often relatively easy and inexpensive to obtain.
In fact, SSL methods are particularly useful in applications where the cost of labeling data is especially expensive, such as medical analysis, natural language processing (NLP), or speech recognition. A subset of SSL methods that have achieved great success in various domains involves algorithms that integrate graph-based techniques. These procedures are popular due to the vast amount of information provided by the graphical framework and the versatility of their applications.
In this work, we propose an algebraic topology-based semi-supervised method called persistent Laplacian-enhanced graph MBO (PL-MBO) by integrating persistent spectral graph theory with the classical Merriman-Bence-Osher (MBO) scheme. Specifically, we use a filtration procedure to generate a sequence of chain complexes and associated families of simplicial complexes, from which we construct a family of persistent Laplacians. Overall, it is a very efficient procedure that requires much less labeled data to perform well compared to many ML techniques, and it can be adapted for both small and large datasets.
We evaluate the performance of the proposed method on data classification, and the results indicate that the proposed technique outperforms other existing semi-supervised algorithms.
}

{\it Keywords}:
Topology-based framework, graph MBO technique, persistent Laplacian, scarcely labeled data


\section{Introduction}

Machine learning has had tremendous success in science, engineering, and many other fields. However, most supervised machine learning algorithms require large amounts of labeled data. Obtaining sufficient amounts of labeled data can be challenging as it is often expensive and time-consuming, and sometimes requires experts in the field. On the other hand, unlabeled data is often available in abundance. Therefore, semi-supervised learning (SSL) \cite{SSL}, which utilizes mostly unlabeled data and much less labeled data for training, has garnered significant attention in the machine learning community.
In particular, one class of semi-supervised learning algorithms that has gained popularity is graph-based semi-supervised learning. The key goal of such methods is to use a graph structure, which is often similarity-based, and both labeled and unlabeled data points, for machine learning tasks.
Here, a graph is often constructed with nodes and edges, where the nodes represent the labeled and unlabeled data set elements, and the edges contain weights that encode the similarity between pairs of data elements.

Overall, graph-based methods, such as those detailed in Section 2.1, have shown great promise for several reasons. First, a similarity graph-based framework provides valuable information about the extent of similarity between data elements through a weighted similarity graph, which is crucial for applications such as data classification and image segmentation. Additionally, graph-based methods yield information about the overall structure of the data.
Second, graph-based methods are capable of incorporating diverse types of data, such as social networks, sensor networks, and biological interaction networks, using a graph structure. This flexibility makes them some of the most competitive methods across a wide range of applications.
In addition, most real-world datasets exist in high-dimensional Euclidean space, but embedding the features into a graphical setting reduces the dimensionality of the problem.

In addition, topological data analysis (TDA) has recently emerged as a powerful tool for analyzing complex data. The central technique of TDA is persistent homology (PH), which combines classical homology and geometric filtration to capture topological changes at different scales. However, PH has some limitations. Specifically, PH fails to capture the homotopic shape evolution of data during filtration.
To overcome this limitation, the authors of \cite{wang} introduce the concept of persistent spectral graphs, also known as persistent combinatorial Laplacians or persistent Laplacians (PLs). In particular, PLs are an extension of the standard combinatorial Laplacian to the filtration setting. It turns out that the harmonic spectra of the PLs return all topological invariants, and the non-harmonic spectra of the PLs provide information about the homotopic shape evolution of the data during filtration.

Motivated by the success of graph-based methods, persistent spectral graph theory, and semi-supervised learning, we propose a novel graph-based method for data classification with low label rates by integrating similarity graph-based threshold dynamics with a family of persistent Laplacians and semi-supervised techniques. The proposed method, called persistent Laplacian-enhanced graph MBO (PL-MBO), adapts the classical MBO scheme developed in \cite{originalMBO} to an LP-based graph framework for data classification. We validate our method using five benchmark classification datasets.

Our contributions are summarized as follows:
\begin{itemize}
    \item  The proposed algorithm uses a family of persistent Laplacian matrices to obtain topological features of data sets that persist across multiple scales. 
    \item The proposed method requires a reduced amount of labeled data for accurate classification compared to many other machine learning methods. It works well even with very low amounts of labeled data, which is important due to the scarcity of labeled data.
    \item The proposed algorithm is very efficient. 
    \item The proposed method can be adapted for both small and large data sets, as outlined in Section 3.2, and works well for both types of data.
\end{itemize}

The remainder of this paper is organized as follows: In Section \ref{Background}, we first  presents the background information on related work,  the graph  framework, and persistent Laplacians. In Section \ref{Method}, we present the graph MBO  technique and   derive our proposed method. The results of the experiments on various benchmark data sets and the discussion of the results are presented in Section \ref{Results}. Finally, Section \ref{Conclusion} provides concluding remarks.

 \section{Background} \label{Background}

\subsection{Related work}

In this section, we will review recent graph-based methods and persistent Laplacian-related algorithms. In particular, graph-based methods usually utilize one of two settings: the transductive setting and the inductive setting. The goal of the transductive setting is to predict the class of elements in the testing set, while the goal of the inductive setting is to learn a function that can predict the class of any data element.

Some of the earliest methods for label inference are label propagation-based methods \cite{LP}. Some general label propagation regularization-based methods include directed regularization \cite{DR}, manifold regularization \cite{MR1, MR2}, anchor graph regularization \cite{AGR}, label propagation algorithm via deformed graph Laplacians \cite{DGL}, Poisson learning \cite{PL}, Tikhonov regularization \cite{TR}, and interpolated regularization \cite{IR}.

Some recently developed label propagation methods involve adaptations of the original Merriman-Bence-Osher (MBO) scheme \cite{originalMBO}, which is an efficient numerical algorithm for approximating motion by mean curvature, to different tasks. In particular, \cite{Merkurjev2} introduces a graphical MBO-scheme based algorithm for segmentation and image processing, while \cite{G-C, Merkurjev1} present a fast multiclass segmentation technique using diffuse interface methods on graphs. Moreover, \cite{meng,sunu} develop algorithms for hyperspectral imagery and video, while \cite{clouds} presents a new graph-based method for the unsupervised classification of 3D point clouds, and \cite{pagerank} incorporates heat kernel pagerank and variations of the MBO scheme. Additionally, \cite{auction} develops a new auction dynamics framework for data classification, which is able to integrate class size information and volume constraints. Furthermore, \cite{MLL} introduces a multiscale graph-based MBO scheme that incorporates multiscale graph Laplacians and adaptations of the classical MBO scheme \cite{originalMBO}. Lastly, \cite{Nicole} develops three graph-based methods for the prediction of scarcely labeled molecular data by integrating transformer and autoencoder techniques with the classical MBO procedure.
 
Other popular semi-supervised methods include shallow graph embedding algorithms. Shallow graph embedding techniques include factorization-based algorithms such as locally linear embedding \cite{LLE}, Laplacian eigenmaps \cite{LE}, the graph factorization algorithm \cite{GF}, GraRep \cite{GraRep}, and HOPE \cite{HOPE}. Moreover, a subset of graph embedding techniques includes those which incorporate random walks; some random walk-based algorithms include DeepWalk \cite{DeepWalk}, Planetoid \cite{Planetoid}, Node2Vec \cite{NV}, LINE \cite{LINE}, and HARP \cite{HARP}.

The success of convolutional neural networks (CNNs) has led to many adaptations of CNNs for graph-based frameworks. In particular, Kipf and Welling's work \cite{GCN} proposes a graph convolutional network (GCN), where the convolutional architecture is developed via a localized first-order approximation of spectral graph convolutions. Additionally, \cite{LI} proposes an adaptive graph convolutional network by constructing a residual graph using a learnable distance function with two-node features as input. In recent years, variants of graph neural networks (GNNs), such as the graph attention network described in \cite{GAN1, GAN2, GAN3}, have been developed and shown great success in deep learning tasks.

In addition to the above, it is important to note that the proposed method utilizes topological tools from persistent spectral graph theory. In particular, while persistent homology is a powerful tool in topological data analysis (TDA) for investigating the structure of data \cite{edelsbrunner2008persistent, zomorodian2004computing}, it is incapable of describing the homotopic shape evolution of data during filtration. Therefore, in \cite{wang}, the authors introduced persistent spectral theory to extract rich topological and spectral information of data through a filtration process, and the theoretical properties of persistent Laplacians are presented in \cite{PLT}. Overall, persistent Laplacians have had tremendous success in computational biology and biophysics, such as protein-ligand binding \cite{PLB}, protein-protein binding problems \cite{chen2022persistent}, and protein thermal stability \cite{wang}. Motivated by the success of topological persistence, we integrate persistent spectral tools to develop our proposed method. Specifically, in this paper, we integrate adaptations of the MBO scheme with persistent Laplacian techniques to develop a new semi-supervised graph-based method for data classification that can achieve good accuracies in cases of data with few labeled elements.


\subsection{Graph-based framework}
\label{graph}

In this section, we review the graph-based framework used in this paper. In particular, let $G = (V, E)$ be an undirected graph, where $V$ and $E$ are the sets of vertices and edges, respectively.   The vertex set $V = \{x_{1}, ... , x_{N}\}$ is associated with the elements of the data set and $E$ is the set of edges connecting pairs of vertices. Importantly, the similarity between vertices $x_i$ and $x_j$ is measured by a weight function $w: V \times V \rightarrow  \mathbb{R}$. The weight function values are usually in the interval $[0,1]$ and are equipped with the following property: a large value of $w(x_{i},x_{j})$ indicates that vertices $x_{i}$ and $x_{j}$ are similar to each other, whereas a small value indicates they are dissimilar; thus, the graph-based framework is able to provide crucial information about the data. The weight function also satisfies the symmetric property. 

While different weight functions have been used in the literature, an often used similarity function is the Gaussian similarity function 
\begin{equation}
\label{gaussian}
w(x_i,x_j) = \exp\left(-\frac{d(x_{i},x_{j})^2}{\sigma ^2}\right),
\end{equation}
where $d(x_{i},x_{j})$ represents a distance (computed using a measure) between vertices $x_{i}$ and $x_{j}$, associated with the $i^{th}$ and $j^{th}$ data elements, and $\sigma\hspace{-0.05cm}>\hspace{-0.05cm}0$ is a parameter which controls scaling in the weight function.

Overall, there are a few important terms to define in a graph-based framework. For example, the degree of vertex $x_{i} \in V$ is defined as
$$d(x_{i}) = \sum_{j} w(x_{i}, x_{j}).$$

Moreover, denote $\mathbf{W}$ as the weight matrix $\mathbf{W}_{i,j} = w(x_{i}, x_{j})$. If $\mathbf{D}$ is the diagonal matrix with the degrees of the vertices as elements, then we can define the graph Laplacian as $\mathbf{L = D - W}$.
In some cases, the graph Laplacian is normalized to account for the behavior that arises when the sample size is large. One example of a normalized graph Laplacian is the symmetric graph Laplacian defined as:
\begin{equation}
\label{symmetric}
\mathbf{L_s = I-D^{-1/2} W D^{-1/2}}.
\end{equation}

In this work, to derive our proposed method, we use a   filtration procedure to generate a sequence of chain complexes and associated families of simplicial complexes and chain complexes, from which we construct a family of persistent Laplacians, which allows us to capture important information from the data. In the next section, we review some background on persistent Laplacians.

\subsection{Persistent Laplacians}
\label{Persistent Laplacians}
In this section, we briefly review some basic notions and definitions to formulate the persistent Laplacian matrix on the simplicial complex. The details can be found in \cite{wang}.

\subsubsection{Simplicial complex}

For $q \geq 0$, a $q-$simplex $\sigma_q$ in an Euclidean space $\mathbb{R}^n$ is the convex hall of a set $P$ of $q+1$ affinely independent points in $\mathbb{R}^n$. In particular,  a 0-simplex is a vertex, a 1-simplex is an edge,  a 2-simplex is a triangle, and a 3-simplex is a tetrahedron.  A $q$-simplex is said to have dimension $q$.

Moreover, a simplicial complex $K$ is a (finite) collection of simplices in $\mathbb{R}^n$ such that
\begin{enumerate}
\item Every face $(\sigma_p)$ of a simplex of $K$ is in $K$.
\item The non-empty intersection of any two simplices of $K$ is a face of each.
\end{enumerate}

\subsubsection{Chain complex}
Let $K$ be a simplicial complex of dimension $q$.  A $q$-chain is a formal sum of $q$-simplices added with some coefficients. Under the addition operation of $\mathbf{Z}_2$, a set of all $q$-chains forms a chain group $C_q(K)$. One can also relate chain groups at different dimensions by a boundary operator: given a $q$-simplex $\sigma_q = \{v_0, \cdots, v_q\}$,  we define the boundary operator $\partial_q^{K}: C_q(K) \rightarrow C_{q-1}(K)$ by
\begin{equation}
\partial_q \sigma_q = \sum_{i =0}^{q} \{v_0, \cdots,\hat{v_i}, \cdots, v_q\},
\end{equation}
where $\hat{v_i}$ indicates the vertex $v_i$ is omitted.  In general, the boundary operator changes a $q$-simplex to a $(q-1)$-simplex.  

A chain complex is a sequence of chain groups connected by   boundary operators.  Similar to boundaries of chains, we have the notion of coboundaries of cochains defined as 
\begin{equation}
    \partial_q^*: C_{q-1}(K) \rightarrow C_q(K).
\end{equation}
Moreover, the $q$-th homology group of $K$ is $ H_q(K) = Z_q (K)/B_q(K)$, where $Z_q (K)$ is $ ker(\partial_q^K)$ and $B_q(K)$ is $im(\partial_{q+1}^{K})$. In addition, the $q$-th Betti number is the rank of the $q$-dimensional homology: $\beta_q^K = rank(H_q(K))$; the Betti number reveals the intrinsic topological information of a geometry or data. Specifically,  $\beta_{0}^t$ provides the number of connected components in $K_t$,  $\beta_{1}^t$ provides the number of one-dimensional circles and  $\beta_{2}^t$ gives the number of two-dimensional voids in $K_t$.
For $K$ oriented simplicial complex,  for $q \geq 0$,  the $q$-combinatorial Laplacian is a linear operator that maps $C_q(K)$ to  $C_q(K)$:
\begin{equation}
\Delta_q := \partial_{q+1} \partial_{q+1}^* + \partial_{q}^* \partial_q.
\end{equation}

Similarly,  we can represent the $q$-combinatorial Laplacian matrix as
\begin{equation}
\mathcal{L}_q = \mathcal{B}_{q+1}\mathcal{B}_{q+1}^T + \mathcal{B}_{q}^T \mathcal{B}_{q},
\end{equation}
where $\mathcal{B}_q$ and $\mathcal{B}_q^T$  is the matrix representation of the $q$-boundary operator and the $q$-coboundary operator $\partial_q^*: C_{q-1}(K) \rightarrow C_q{K}$ defined in \cite{book}, respectively. Note that when $K$ is a graph, we have $ \mathcal{L}_0(K) = \mathcal{B}_{1}\mathcal{B}_{1}^T + \mathcal{B}_{0}^T \mathcal{B}_{0}$.  This means that the $0$-combinatorial Laplacian matrix is similar to  a graph Laplacian matrix defined in the Section \ref{graph} except that their matrix  values are very different.
Note that the use of simplicial complexes allows high-dimensional modeling in combinatorial Laplacians, whereas graph Laplacians can only support pairwise interactions.

\subsubsection{Filtration }
The notion of filtration  is at the core of topological persistence. In particular, a filtration can be defined in the context of topological spaces or simplicial complexes. Specifically, a filtration of $\mathcal{F}=\mathcal{F}(K)$ of a oriented simplicial complex $K$ is a nested sequence of its subcomplexes \\
\begin{equation}
\mathcal{F} : \phi = K_0 \subseteq K_1 \subseteq \cdots \subseteq K_n = K.
\end{equation}
Overall, it induces a sequence of chain complexes:

\begin{equation}
\left.\begin{array}{cccccccccccccc}
\cdots & C_{q+1}^1 &
\xrightleftharpoons[\partial_{q+1}^{1^\ast}]{\partial_{q+1}^1} & C_q^1 &
\xrightleftharpoons[\partial_q^{1^\ast}]{\partial_q^1} & \cdots & \xrightleftharpoons[\partial_3^{1^\ast}]{\partial_3^1} & C_2^1 & \xrightleftharpoons[\partial_2^{1^\ast}]{\partial_2^1} & C_1^1 & \xrightleftharpoons[\partial_1^{1^\ast}]{\partial_1^1} & C_0^1 & \xrightleftharpoons[\partial_0^{1^\ast}]{\partial_0^1} & C_{-1}^1 \\
& \rotatebox{-90}{$\subseteq$} &  & \rotatebox{-90}{$\subseteq$} &  &  &  & \rotatebox{-90}{$\subseteq$} &  & \rotatebox{-90}{$\subseteq$} &  & \rotatebox{-90}{$\subseteq$} &  &  \\
\cdots & C_{q+1}^2 &
\xrightleftharpoons[\partial_{q+1}^{2^\ast}]{\partial_{q+1}^2} & C_q^2 &
\xrightleftharpoons[\partial_q^{2^\ast}]{\partial_q^2} & \cdots &
\xrightleftharpoons[\partial_3^{2^\ast}]{\partial_3^2} & C_2^2 & \xrightleftharpoons[\partial_2^{2^\ast}]{\partial_2^2} & C_1^2 & \xrightleftharpoons[\partial_1^{2^\ast}]{\partial_1^2} & C_0^2 & \xrightleftharpoons[\partial_0^{2^\ast}]{\partial_0^2} & C_{-1}^1 \\
& \rotatebox{-90}{$\subseteq$} &  & \rotatebox{-90}{$\subseteq$} &  &  &  & \rotatebox{-90}{$\subseteq$} &  & \rotatebox{-90}{$
	\subseteq$} &  & \rotatebox{-90}{$\subseteq$} &  &  \\
& \rotatebox{-90}{$\cdots$} &  & \rotatebox{-90}{$\cdots$} &  &  &  & \rotatebox{-90}{$\cdots$} &  & \rotatebox{-90}{$\cdots$} &  & \rotatebox{-90}{$\cdots$} &  &  \\
& \rotatebox{-90}{$\subseteq$} &  & \rotatebox{-90}{$\subseteq$} &  &  &  & \rotatebox{-90}{$\subseteq$} &  & \rotatebox{-90}{$
	\subseteq$} &  & \rotatebox{-90}{$\subseteq$} &  &  \\
\cdots & C_{q+1}^m &
\xrightleftharpoons[\partial_{q+1}^{m^\ast}]{\partial_{q+1}^m} & C_q^m &
\xrightleftharpoons[\partial_q^{m^\ast}]{\partial_q^m} & \cdots &
\xrightleftharpoons[\partial_3^{m^\ast}]{\partial_3^m} & C_2^m & \xrightleftharpoons[\partial_2^{m^\ast}]{\partial_2^m} & C_1^m & \xrightleftharpoons[\partial_1^{m^\ast}]{\partial_1^m} & C_0^m & \xrightleftharpoons[\partial_0^{m^\ast}]{\partial_0^m} & C_{-1}^1,
\end{array}\right.
\end{equation}

\vspace{0.1cm}
where $C_{q}^{t}:= C_q(K_t)$ and $\partial_{q}^{t}: C_{q}(K_t) \rightarrow C_{q-1}(K_t)$. 

\vspace{0.2cm}

\subsubsection{Persistent Laplacians}
\label{persistent}

At the core of our proposed method is the persistent Laplacian matrix. In this section, we discuss the concept in more detail. Consider $\mathbb{C}_{q}^{t+p}$, the subset of $C_{q}^{t+p}$ whose boundary is in $C_{q-1}^t$, defined by:
\begin{equation}
\mathbb{C}_{q}^{t+p} = \{e \in C_{q}^{t+p} | \partial_{q}^{t+p} (e) \in C_{q-1}^{t}\} \subseteq C_{q}^{t+p}.
\end{equation}
We define the $p$-persistent $q$-boundary operator as $\cpartial_{q}^{t+p}: \mathbb{C}_{q}^{t+p} \rightarrow C_{q-1}^{t}$ and the adjoint boundary operator as $(\cpartial_{q}^{t+p})^*: C_{q-1}^t \rightarrow \mathbb{C}_{q}^{t+p}$; both operators are well-defined. The $p$-persistent $q$-combinatorial Laplacian operator is defined as:
$$\Delta_{q}^{t+p} = \cpartial_{q+1}^{t+p} (\cpartial_{q+1}^{t+p})^* + (\partial_{q}^t)^* \partial_q^t.$$
Now, we denote the matrix representation of $\cpartial_{q+1}^{t+p}$ and $\cpartial_{q}^{t} $ by $\mathcal{B}_{q+1}^{t+p}$
and $\mathcal{B}_{q}^{t}$, respectively. Similarly, we can represent $(\cpartial_{q+1}^{t+p})^*$ and $(\cpartial_{q}^t)^*$ by matrices $(
\mathcal{B}_{q+1}^{t+p})^T$ and $(\mathcal{B}_{q}^{t})^T$, respectively. Therefore, the $p$-persistent $q$-combinatorial Laplacian matrix  is defined as:
\begin{equation}
\mathcal{L}_{q}^{t+p} = \mathcal{B}_{q+1}^{t+p}(\mathcal{B}_{q+1}^{t+p})^T + (\mathcal{B}_{q}^{t})^T \mathcal{B}_{q}^{t}.
\end{equation}
We also denote the set of spectral of $\mathcal{L}_{q}^{t+p}$ by
$$\text{Spectra}(\mathcal{L}_{q}^{t+p}) = \{(\lambda_1)_{q}^{t+p}, (\lambda_2)_{q}^{t+p}, \cdots, (\lambda_N)_{q}^{t+p}\},$$
where the spectra are arranged in ascending order. Moreover, the $p$-persistent $q$th Betti numbers can be defined as the number of zero eigenvalues of the $p$-persistent $q$-combinatorial Laplacian matrix $\mathcal{L}_{q}^{t+p}$. Thus,
   $$ \beta_{q}^{t+p} = \text{dim}(\mathcal{L}_{q}^{t+p})-\text{rank}(\mathcal{L}_{q}^{t+p}) = \text{nullity}(\mathcal{L}_{q}^{t+p}) = \# \text{of zero eigenvalues of} \mathcal{L}_{q}^{t+p}.$$

\begin{figure}
    \centering
    \includegraphics[width=17cm, height=10cm]{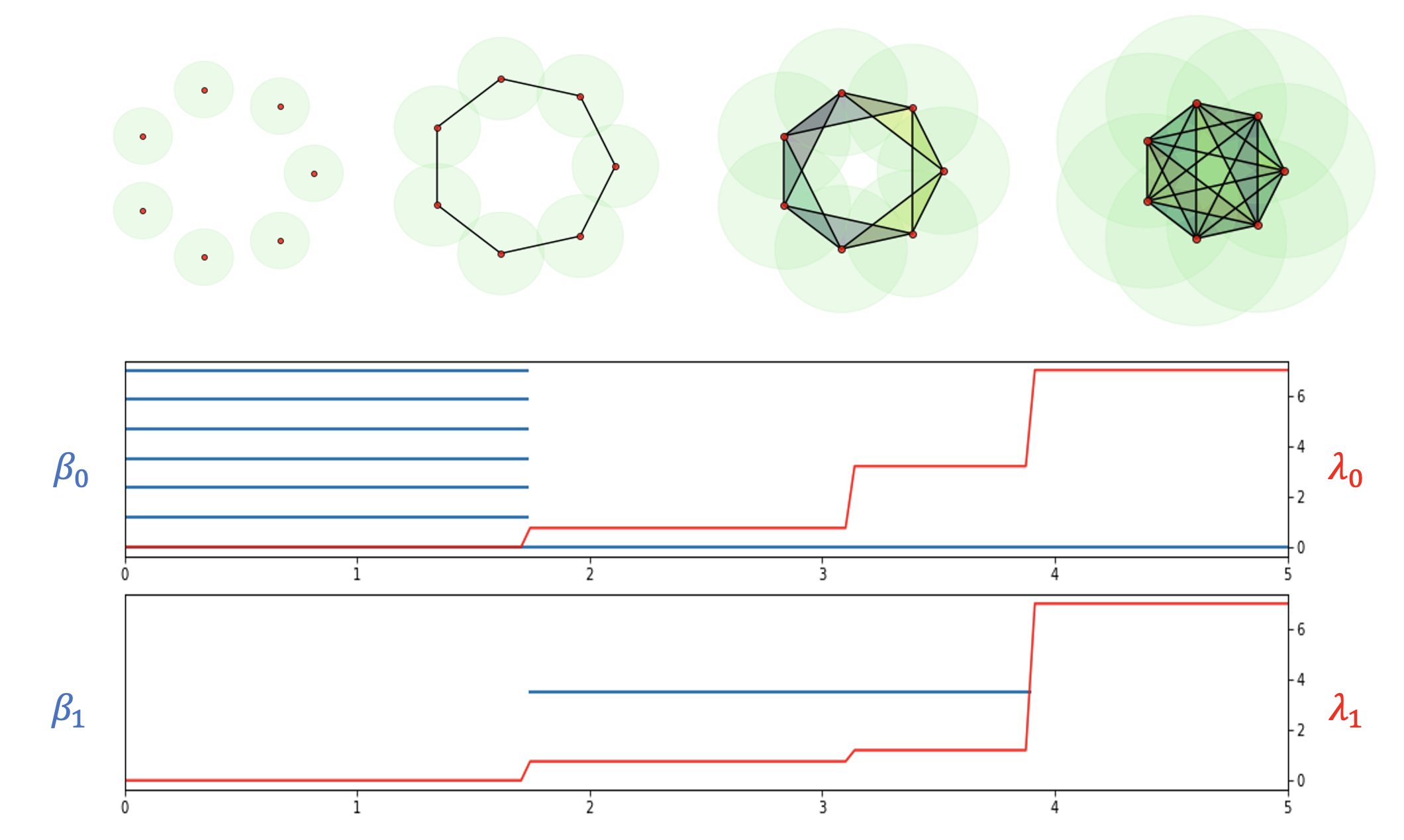}
    \caption{Comparison of persistent homology and persistent Laplacians. The top panel shows the same data (seven points) at four different   stages of  filtration characterized  by a radius $r$. The second and third panels represent the corresponding topological (i.e., harmonic) and non-harmonic features of dimension 0 and dimension 1, respectively. 
The blue line represents the persistent homology (PH) barcodes of dimension 0 ($\beta_{0}(r)$) and dimension 1 ($\beta_{1}(r)$), while the red line represents the first non-zero eigenvalues of dimension 0 $(\lambda_{0}(r))$ and dimension 1 $(\lambda_{1}(r))$ of the persistent Laplacians (PLs). It is shown that the harmonic spectra of PLs return all topological invariants of PH and the non-harmonic spectra of PLs reveal the additional homotopic shape evolution of PLs during the filtration (i.e., the second jumps in the red curves correspond to the increase in connectivity, the third geometric shape, but there was no topological change). Note that PH fails to capture the homotopic shape evolution during filtration. The details on persistent spectral graphs can be found in \cite{wang}.}
    \label{filtration}
    \vspace{0.3cm}
\end{figure}

In the above definition, $ \beta_{q}^{t+p}$ counts the number of $q$-cycles in $K_t$ that are still alive in $K_{t+p}$. This topological information is exactly what can be obtained using persistent homology. However, persistent spectral theory provides additional geometric information using the spectra of persistent combinatorial Laplacians. More specifically, the non-harmonic spectra can capture both the topological changes and the homotopic shape evaluation of the data; see Figure \ref{filtration} for an illustration. More detailed descriptions of persistent spectral graphs can be found in the work \cite{wang}. 

\section{Methods  } \label{Method}


This section provides the  graph MBO technique and its LP generalization.

\subsection{The graph MBO technique}
\label{graphMBO}

The proposed data classification method derived in this paper is based on the techniques outlined in the literature, such as \cite{G-C, Merkurjev1, Merkurjev2, Nicole, MLL}, which generalize the original MBO algorithm \cite{originalMBO} into a graphical setting. 

For the derivation of our proposed method, consider a matrix $\mathbf{U} = (u_1,  \ldots, u_N)^T \in \mathbb{R}^{N, K}$, where $K$ is the number of classes, $N$ is the number of data elements and $u_{i} \in \mathbb{R}^K $ indicates the probability distribution over the different classes for the data element $x_{i}$; thus, the $j^{th}$ element of $u_i$ is the probability of data element $x_i$ of belonging to class $j$. In particular, the vector $u_{i}$ is an element of the Gibbs simplex $\Sigma ^{K}$ defined as:
\begin{equation}
\Sigma^{K} :=\{ (x_{1}, \ldots, x_{K}) \in [0,1]^K |  \sum_{k =1}^{K} x_{k} = 1 \}.
 \label{gibbs}
 \end{equation}
 The goal of the proposed technique will be to compute the optimal matrix $\mathbf{U}$.

Regarding the data classification problem of machine learning, many data classification algorithms can be regarded as optimization techniques where a specific objective function is optimized. In particular, one of the popular optimization techniques for data classification involves minimizing the following general objective function:
\begin{equation}
E(\mathbf{U}) = R(\mathbf{U}) + {  F }(\mathbf{U}),
\end{equation}
where $\mathbf{U}$ is a classification variable with each row representing a probability distribution of a particular data element over the classes,  $R(\mathbf{U})$ is a regularization term and ${  F }(\mathbf{U})$ is a fidelity term containing information about labeled data. The goal is to minimize $E(\mathbf{U})$ and thus to obtain the optimal $\mathbf{U}$. 


Regarding the regularization term $R(\mathbf{U})$, it may be useful to examine a certain functional called the Ginburg-Landau (GL) functional, described in more detail in \cite{Merkurjev2}. The classical GL functional takes the following form:
\vspace{-0.1cm}
\begin{equation}
\label{GL}
{\rm GL}(u) = \frac{\epsilon}{2} \int |\nabla u|^{2} dx + \frac{1}{\epsilon} \int W(u) dx,
\end{equation}
where $u$ is a scalar field representing the state of the phases in the system,  $W(u)$ is a double-well potential, $\epsilon$ is a positive constant and $ \nabla$ denotes the spatial gradient operator. In \cite{Bertozzi1, G-C}, the authors modify the original Ginzburg-Landau functional (\ref{GL}) into a graph-based functional by replacing the first term with the graph Dirichlet energy to obtain a graph-based regularization term. In addition, to incorporate multiple classes, they modify the double-well potential into a multi-class setting; one can also add an $\mathbf{L}^{2}$ penalty term to incorporate the labels of the labeled elements. For more details about this graph-based representation of the Ginzburg-Landau functional, one can refer to \cite{  G-C, Merkurjev1, Merkurjev2}. 

Inspired by the above, the optimization problem we consider in this work consists of minimizing the following graph-based Ginzburg-Landau energy:
\vspace{-0.125cm}
\begin{equation}
\label{Energy}
E(\mathbf{U}) = \frac{\epsilon}{2} \langle \mathbf{U}, \mathbf{L_{norm}U} \rangle + \frac{1}{2\epsilon}\sum_{i \in V}(\prod_{k=1}^K \frac{1}{4}||u_{i} - e_{k}||_{L_1}^2) + \sum_{i \in V} \frac{\mu_{i}}{2}||u_{i} - \hat{u}||^2,
\end{equation}
where $\langle \mathbf{U, L_{norm}U}\rangle$ = trace $(\mathbf{U}^T \mathbf{L_{norm}U})$,  $\mathbf{L_{norm}}$ is any normalized graph Laplacian such as the symmetric graph Laplacian, $K$ is number of classes, and $u_{i}$ is the $i^{th}$ row of $\mathbf{U}.$  In addition, $\hat{u_{i}}$ is a vector indicating the prior class knowledge of $x_{i}$, $e_{k}$ is an indicator vector of size $K$ with a one in $k^{th}$ component and zero elsewhere, and $\mu_{i}$ takes some positive value if $x_{i}$ is labeled data element and 0 otherwise.

To minimize the graph-based multiclass energy functional (\ref{Energy}), the authors of \cite{G-C} developed a convex splitting scheme. Similarly, the authors of \cite{Merkurjev2} derived a modified MBO scheme to minimize (\ref{Energy}). More specifically, the authors drew upon a technique that can be used to minimize the classical non-graphical GL functional (\ref{GL}), which can be optimized in the $\mathbf{L}_{2}$ sense using gradient descent, resulting in an Allen-Cahn equation. If a time-splitting scheme is then applied, one obtains a procedure where one alternates between propagation using the heat equation with a forcing term and thresholding, which is similar to the steps of the original MBO scheme \cite{originalMBO}, an efficient numerical technique for computing mean curvature flow. This can be extended to a graphical and multiclass setting by using a graph Laplacian and projecting to the closest vertex in the Gibbs simplex \eqref{gibbs}. For a detailed explanation, one can refer to \cite{Merkurjev2}.

Overall, the goal of this paper is to integrate similar techniques with persistent spectral graphs and a  filtration procedure,  from which we construct a family of persistent Laplacians. Incorporating persistent Laplacians into our proposed procedure will allow us to obtain crucial information about the data, such as all topological invariants and the homotopic shape evolution of the data during the filtration.

\subsection{PL-MBO algorithm} \label{tpm}

In this section, we will derive our proposed semi-supervised method, PL-MBO, for data classification, which is especially useful for cases with low label rates. To derive the proposed method, the PL-MBO algorithm, we will consider minimizing a variant of \eqref{Energy}, where instead of choosing the graph Laplacian to be the normalized graph Laplacian, we will choose it from a family of persistent Laplacians. 

Consider a simple graph; we can then generate persistent Laplacians from a weighted graph Laplacian by using a threshold in the matrix computation. In particular, we define the weighted Laplacian as $\mathbf{L_{norm}} =(L_{ij}) $, where $L_{ii} = -\sum_{j =1}^{N} L_{ij}$ and $L_{ij} \leq 0$ for all $i$ and $j$, and $N$ is the number of data elements in the particular data set.

For  $i \neq j,$ let $L_{\rm max}$ = $\max\limits_{ij} L_{ij}$, $L_{\rm min} = \min\limits_{ij} L_{ij}$, and $d = L_{\rm max} - L_{\rm min}.$

 Let $L_n$ be an integer greater than $1$. We can then define the $k${th} persistent Laplacian, $\mathbf{L}^k_{\rm persistent} $, for ${k} = 1,2, \cdots, L_n$, as $(\mathbf{L}^k_{\rm persistent})_{ij} = L_{ij}^{k}$, where, for $i \neq j$:
\begin{equation}
\label{PL}
 L_{ij}^{k} = \begin{cases}
    0 &  $ if  $ L_{ij} \le \frac{{ k}}{L_n}  d + L_{\rm min}, \\
        -1 & \text{otherwise}.
    \end{cases}
\end{equation}
The diagonal entries of the persistent Laplacian, $\mathbf{L}^k_{\rm persistent}$, are computed as  
\begin{equation}
\label{PL2}
 L_{ii}^{k} = - \sum_{j=1}^{N} L_{ij}^{k}.
 \end{equation}

 
In our method,  $\{\mathbf{L}^k_{\rm persistent}\}$ (for $k = 1,2, \cdots, L_n$) are used in place of $\mathbf{L_{norm}}$ in Eq. (\ref{Energy}). 
Specifically,  a variant of \eqref{Energy} with each derived persistent Laplacian from the family is minimized using similar techniques to those described at the end of Section \ref{graphMBO}. Finally, the results from each persistent Laplacian assisted MBO are concatenated and fed into a classifier, such as a gradient-boosting decision tree, support vector machine or a random forest, which predicts the final class of each data element.

In particular, let $\mathbf{U}$ represent a matrix where each row $u_i$ contains the probability distribution of each data element over the classes. Also, let $dt >0$ be the step size,  $N$ be the number of data elements, and $K$ be the number of classes. Moreover, let the vector \boldsymbol{$\mu$} represent a vector that takes the value $\mu$ at labeled data elements and $0$ at unlabeled data elements. In addition, we define the $\mathbf{U_{labeled}}$ matrix as follows: for labeled elements, each row is an indicator vector with a 1 in the entry corresponding to the class of the labeled element. All other entries are set to $0$.    
 
We can summarize our proposed method as follows:
\vspace{0.25cm}
\begin{itemize}
  \setlength\itemsep{0.4em}
   \item Using the input data, construct a similarity graph using a chosen similarity function such as \eqref{gaussian}, and then compute the symmetric graph Laplacian \eqref{symmetric}. 
   \item Using the symmetric graph Laplacian, construct a family of $L_n$ persistent Laplacians, i.e. $\{\mathbf{L}^k_{\rm persistent}\}$, for $k = 1,2, \cdots, L_n$, where $L_n$ is an integer greater than one, as derived in Section \ref{persistent}.
   \item For each persistent Laplacian in the family, i.e. $\mathbf{L}^k_{\rm persistent}$, for $k = 1,2, \cdots, L_n$, compute the smallest $N_e$ eigenvalues and associated eigenvectors. Only a small portion of the eigenvalues and corresponding eigenvectors need to be computed.
   \item Initialize $\mathbf{U}$ using techniques such as random initialization or Voronoi initialization. In particular, in Voronoi initialization, the labels of the unlabeled points are initialized by creating a Voronoi diagram with the labels of the labeled points as the seed points; every point is assigned the label of the labeled point in its Voronoi cell. We note that, in any initialization, the rows of the initial $\mathbf{U}$ corresponding to labeled points should consist of indicator vectors with a 1 at the place corresponding to the true class of the data element.
   
   \item For each persistent Laplacian in the family, i.e. $\mathbf{L}^k_{\rm persistent} $, for $k = 1,2, \cdots, L_n$, perform the following MBO-like steps for $N_t$ iterations, to obtain the next iterate of $\mathbf{U}$; if there are $L_n$ persistent Laplacians in the family, there will be $L_n$ output matrices $\mathbf{U}$:
\vspace{0.1cm}
\begin{enumerate}
\item  Heat equation with a forcing term: $$\mathbf{U}^{n+\frac{1}{2}} =\mathbf{U}^{n} - dt \{\mathbf{L}^k_{\rm persistent} \mathbf{U}^{n+\frac{1}{2}}+\boldsymbol{\mu} \cdot (\mathbf{U}^{n}-\mathbf{U_{labeled}})\},$$ where $\boldsymbol{\mu}$ is a vector which takes a value $\mu$ in the $i^{th}$ place if $\mathbf{x}_i$ is a labeled element and $0$ otherwise, and the term $\boldsymbol{\mu} \cdot (\mathbf{U}^{n}-\mathbf{U_{labeled}})$ indicates row-wise multiplication by a scalar. Later, we describe the spectral techniques used to make the first step efficient even for larger data sets.
\vspace{0.3cm}
\item Projection to simplex: Each row of $\mathbf{U}^{n+\frac{1}{2}}$ is projected onto the simplex using \cite{chen}.
\vspace{0.25cm}
\item Displacement: $ u_i^{n+1}= \boldsymbol{e}_k$,
where $u_i^{n+1}$ is the $i^{th}$ row of $\mathbf{U}^{n+1}$, and $\boldsymbol{e}_k$ is the indicator vector. 
\end{enumerate}
\vspace{0.25cm}
\item Concatenate the results of each output matrix (from each persistent Laplacian) to form a new matrix. For the binary case, one only needs to concatenate the first column of each output matrix to form the new matrix.
\item Divide the new matrix into training data and testing data. The rows corresponding to $\boldsymbol{ \mu_i} = \mu$ will be used for training and the rows corresponding to $\boldsymbol{ \mu_i} = 0$ will be used for testing.  
\item Use a classifier, such as a gradient-boosting decision tree, a support vector machine or a random forest, to predict the final class of the unlabeled data elements.
\end{itemize} 

\vspace{0.5cm}
 
\begin{figure}
    \centering
    \includegraphics[width=14cm, height=4cm]{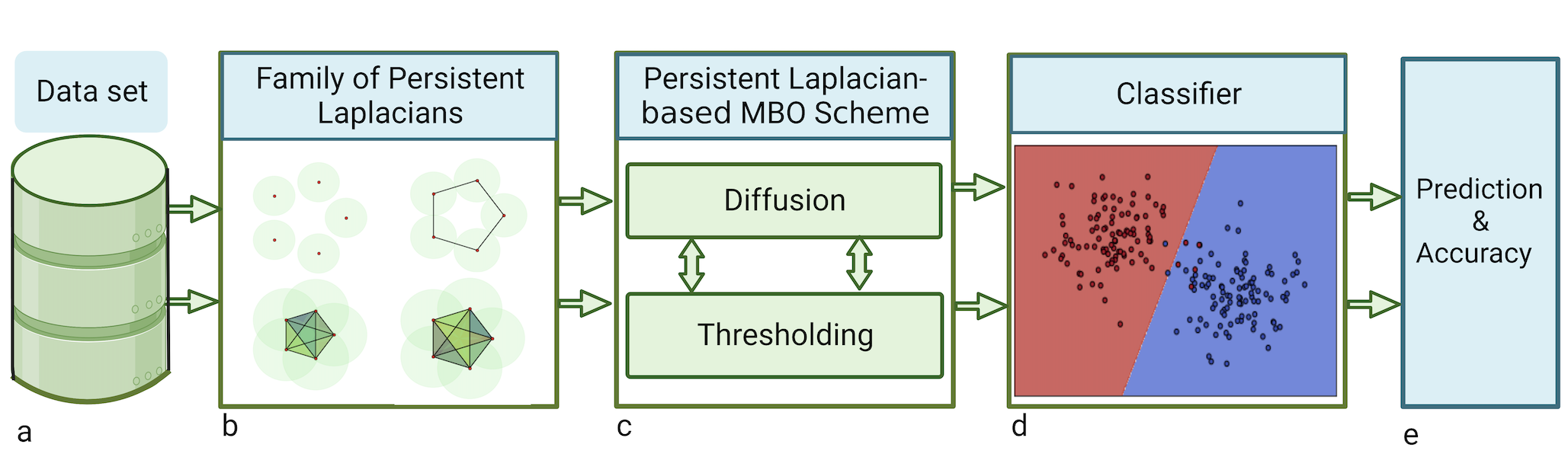}
    \caption{Visual description of PL-MBO. (a) From a given data set, a similarity graph is constructed using a chosen similarity function. (b) A family of persistent Laplacians is formed as derived in Section \ref{Persistent Laplacians}. (c) The graph-based Ginzburg-landau energy (\ref{Energy}) is minimized using a modified persistent Laplacian-based MBO scheme incorporating the family of persistent Laplacians; a new test set is formed using the output. (d) A machine learning algorithm is used to classify the new test data. (e) The accuracy of the proposed method is computed on the test data set. }
    \label{method visualization}
    \vspace{0.6cm}
\end{figure}

The proposed PL-MBO procedure is detailed as Algorithm \ref{alg-pl-mbo}. For an illustration, an intuitive interpretation of the proposed method is shown in Figure \ref{method visualization}.

To make the scheme even more efficient, we use a spectral technique and utilize a low-dimensional subspace spanned by only a small number of eigenfunctions, similar to the procedures outlined in \cite{Merkurjev1, G-C, Merkurjev2}. The idea is to first rewrite Step 1 (Heat equation with a forcing term) as
\begin{equation}
\label{step1}
\mathbf{U}^{n+\frac{1}{2}} = \mathbf{C}^{-1}\mathbf{U_{update}},
\end{equation}
where  
\begin{equation}
\mathbf{C} = \mathbf{I} + dt \mathbf{L}^k_{\rm persistent} \quad \text{and} \quad \mathbf{U_{update}}= \mathbf{U^n} - dt \boldsymbol{\mu} \cdot (\mathbf{U}^{n}-\mathbf{U_{labeled}}).
\end{equation}
Now, let the eigendecomposion of $ \mathbf{L}^k_{\rm persistent}$ be denoted as $ \mathbf{L}^k_{\rm persistent}$ = $\mathbf{X}  \mathbf{\Lambda} \mathbf{X^T}$, and let $\mathbf{X_{truncated}}$ and $\mathbf{\Lambda_{truncated}}$ be truncated matrices of $\mathbf{X}$ and $\mathbf{\Lambda}$, respectively, containing only $N_e$ smallest eigenvectors and eigenvalues of the persistent Laplacian, respectively, where $N_e << N$. Therefore, $\mathbf{X_{truncated}}$ and $\mathbf{\Lambda_{truncated}}$ are matrices of size $N \times N_e$ and $N_e \times N_e$; thus, at least one dimension is very small. One can then rewrite \eqref{step1} as
\begin{equation}
\label{step2}
\mathbf{U^{n+\frac{1}{2}}} = \mathbf{X_{truncated}} (\mathbf{I} + dt \mathbf{\Lambda_{truncated}})^{-1} \mathbf{X_{truncated}^T} \mathbf{U_{update}},
\end{equation}
where $\mathbf{U_{update}} =\mathbf{U^n}- dt \boldsymbol{\mu} \cdot (\mathbf{U}^{n}-\mathbf{U_{labeled}}$). Note that all of the aforementioned matrices in \eqref{step2} have at least one dimension which is small, and that only the smallest $N_e$ eigenvalues of the persistent Laplacian and their corresponding eigenvectors need to be computed.

Overall, this spectral technique allows the diffusion operation of the proposed scheme, particularly Step 1, which involves the heat equation with a forcing term, to be decomposed into faster matrix multiplication, resulting in an efficient proposed algorithm. It is also important to note that the graph weights are only used to compute the first $N_e$ eigenvalues. Once they are computed, the main steps of the scheme involve only the truncated matrices and $\mathbf{U}$, which allows the scheme to be very fast.

\begin{algorithm}[H]
\caption{PL-MBO Algorithm }\label{alg-pl-mbo}
\begin{algorithmic}[H]
\vspace{0.075cm}
\Require $N$ (\# of data set elements), $m$ (\# of labeled data elements), labeled data $\mathcal{L}=\{(\mathbf{x}_i, y_i)\}_{i=1}^m$, where $y_i$ is the label of $\mathbf{x}_i$, unlabeled data $\mathcal{U}=\{\mathbf{x}_j\}_{j=m+1}^N$, $N_n$ (\# of nearest neighbors),  $\sigma>0$,  $dt >0$, $N_e << N$ (\# of eigenvectors to be computed), $N_t$ (maximum \# of iterations), $\bm{\mu}$ (an $N \times 1$ vector which takes a value $\mu$ in the $i^{th}$ place if $\mathbf{x}_i$ is a labeled element and $0$ otherwise), $L_n$ (\# of persistent Laplacians).
\vspace{0.1cm}
\Ensure Prediction of the final class for each data element.
\vspace{0.1cm}
\State 1: Construct a $N_n$-nearest neighbors graph from the data.
\vspace{0.1cm}
\State 2: Compute the symmetric graph Laplacian \eqref{symmetric}. 
\vspace{0.1cm}
\State 3: Construct a family of $L_n$ persistent Laplacians from the graph Laplacian using \eqref{PL} and \eqref{PL2}. 
\vspace{0.1cm}
\State 4: Compute the matrices $\mathbf{U_{labeled}}$, $\mathbf{\Lambda_{truncated}}$ and $\mathbf{X_{truncated}}$ for each persistent Laplacian as described in Section \ref{tpm}. Let $\mathbf{X_p}= \mathbf{X_{truncated}}$ for $\mathbf{L^p_{persistent}}$ and $\mathbf{\Lambda_p}= \mathbf{\Lambda_{truncated}}$ for $\mathbf{L^p_{persistent}}$.
\vspace{0.1cm}
\State 5: Complete the following steps starting with $n = 1$:
\vspace{0.3cm}
\For{$i = 1 \to N$ and all $k$}
\vspace{0.1cm}
\State $\mathbf{U}_{i k}^{~0} \leftarrow rand((0,1))$
\vspace{0.1cm}
\State $ \mathbf{u}_i^{0} \leftarrow  projectToSimplex(\mathbf{u}_i^{0})$ using \cite{chen}, where $\mathbf{u}_i^{0}$ is $i^{th}$ row of $\mathbf{U}^0$.
\State $\mathrm{If~} \bm{\mu}_i > 0, ~ \mathbf{U}_{i k}^{~0} \leftarrow \mathbf{{U_{\text{labeled}}}}_{i k}$ 

\EndFor 
\For{$p = 1 \to L_n$}
\State $\mathbf{B_p} \leftarrow \left(\mathbf{I} +{\text{dt}}\bm{\Lambda_p}\right)^{-1}\mathbf{X_p}^T$
\vspace{0.1cm}
\While{$n < N_t$}
\State $\mathbf{F} \leftarrow  \mathbf{U}^n - {\text{dt}} \, \boldsymbol{\mu} \cdot (\mathbf{U}^n-\mathbf{U_\text{labeled}})$
\State $\mathbf{A_p} \leftarrow \mathbf{B_p}\mathbf{F}$
\State $\mathbf{U_p}^{n+1} \leftarrow \mathbf{X_p}\mathbf{A_p}$
\vspace{0.1cm}
\For{$i =1 \to N$} 
\State$ \mathbf{v}_i^{n+1} \leftarrow  projectToSimplex(\mathbf{u}_i^{n+1})$ using \cite{chen}
\State$ \mathbf{u}_i^{n+1} \leftarrow \boldsymbol{e}_k$, where $k$ is closest simplex vertex to $\mathbf{v}_i^{n+1}$
\EndFor
\State The matrix $\mathbf{U_p}^{n+1}$ is such that its $i^{th}$ row is $\mathbf{v}_i^{n+1}$. 
\State $ n \leftarrow n + 1$
\EndWhile
\State $\mathbf{X} =  Concatenate\{\mathbf{U_p}_{p=1}^{L_n}\}$
\EndFor
\State $j, k=1$
\For{$i =1 \to N$} 
\State \textbf{If} ($\bm{\mu}_i > 0)$
\State $\mathbf{X_{train}(j,:) = X(i,:)}$ \text{and} $\mathbf{Y_{train}(j,1) = Y(i,1)}$
\State $ j \leftarrow j + 1$
\EndFor
\For{$i =1 \to N$} 
\State \textbf{If} ($\bm{\mu}_i = 0)$
\State $\mathbf{X_{test}(k,:) = X(i,:)}$ \text{and} $\mathbf{Y_{test}(k,1) = Y(i,1)}$
\State $ k \leftarrow k + 1$
\EndFor
\State 6: Use classifier to predict the final class of $\mathbf{X_{test}}$.

\vspace{0.1cm}
\vspace{0.3cm}
\end{algorithmic}
\end{algorithm}

Regarding the computation of the few eigenvalues and their corresponding eigenvectors (for a particular persistent Laplacian derived in Section \ref{Persistent Laplacians}), which our proposed method requires, we note that there are many methods available for the task. For sparse matrices, such as the family of persistent Laplacians derived in this paper, and moderately small datasets, the authors in \cite{Bertozzi1, G-C} suggest using the Rayleigh-Chebyshev procedure \cite{R-C}. This method is a modified version of the inverse subspace iteration algorithm. For large and fully connected graphs, the Nystr\"{o}m extension technique \cite{NE1, NE2} is recommended. The Nystr\"{o}m extension algorithm is a matrix completion method that incorporates computations using much smaller submatrices of lower dimensions, thus saving computational time, and approximates eigenvectors and eigenvalues using a quadrature rule with randomly chosen interpolation points. Moreover, this technique requires the computation of only a very small portion of the graph weights, making this procedure very efficient even for very large datasets. For simplicity, in our experiments, we use MATLAB's $eigs$ function to compute the $N_e$ smallest eigenvalues along with their corresponding eigenvectors. 


 For small data sets, like some of those used in the experiments of this paper, it is more desirable to compute the graph weights directly by calculating pairwise distances. In this case, the efficiency of the task can be increased by using a parallel computing technique or by reducing the dimension of the data. Then, a graph is often made sparse using, for example, thresholding or an $l$ nearest neighbors technique, resulting in a similarity graph where most of the edge weights are zero. Overall, a nearest neighbor graph can be computed efficiently using the $kd$-tree code of the VLFeat open source library \cite{kd-tree}.

           \vspace{0.5cm}

\section{Results and discussion} \label{Results}
\subsection{Data sets}
We tested our proposed method on five benchmark data sets: two artificial data sets and three real-world data sets. The data sets are as follows:
\vspace{0.15cm}
\begin{itemize}
  \setlength\itemsep{0.4em}
\item The G50C data set \cite{g_50c} is an artificial data set inspired by \cite{g50c}: the data is generated from two standard normal multivariate Gaussians.  This data set contains $550$ points located in a 50-dimensional space such that the Bayes error is $5$\%. There are two classes. Fifty labeled elements are used for the experiments.

\item The WebKB (World Wide Knowledge Base) data set \cite{webkb} contains web pages collected from the computer science departments of four different universities: Cornell University, University of Texas, University of Washington, and University of Wisconsin. For our experiments, we used a subset of the WebKB data set consisting of 1051 sample points that were manually classified into two categories: course and non-course. Each web document is described by the text on the web pages (called page representation) and the anchor text on the hyperlinks pointing to the page (called link representation). The information from the text of each data set element can be encoded with a 4840-component vector. We used twelve labeled elements for our experiments.

\item The Haberman data set \cite{haberman} contains information on the survival of patients who had undergone surgery for breast cancer. The study was conducted between 1958 and 1970 at the University of Chicago's Billings Hospital. It consists of 306 data points located in a 3-dimensional space, with two classes: patients who survived 5 years or longer, and patients who died within 5 years. For our experiments, we used 60 labeled data points.

\item The Madelon data set \cite{madelon} is a synthetic data set consisting of data points grouped into 32 clusters, placed at the vertices of a five-dimensional hypercube, with two classes. The five dimensions constitute five informative features, and 15 linear combinations of those features were added to form a set of 20 redundant informative features. The goal is to classify examples into two classes based on these 20 features. The dataset also contains distractor features called 'probes' with no predictive power. The order of the features and patterns was randomized. Overall, this data set contains 2000 elements, all of which are in a 500-dimensional space, and 200 labeled elements are used for the experiments.

\item The Banana data set \cite{banana} is a two-dimensional binary classification dataset that contains two banana-shaped clusters; thus, there are two classes in this data set. There are 5300 elements in the data set, which each element being 2-dimensional. Fifty labeled elements are used for the experiments.
\end{itemize}

\vspace{0.3cm}

The details of the data sets are outlined in Table \ref{tab: datasets}.

\vspace{0.3cm}

\begin{table*}[!ht]

	\centering
	{
		\caption{Data sets used in the experiments.}
		\label{tab: datasets}
		\vspace{0.05cm}
		\begin{tabular}{lcccc}
			\hline
			Data set   & No. of data elements & Sample dim.  & No. of labeled data  & No. of classes \\
			\hline
                Haberman & 306 & 3 & 60 & 2 \\
			G50C				 &	550         &	50		 & 50 & 2  \\
                WebKB (page+link)   &  1051    &    4840      &      105   &  2 \\
			Madelon       & 2000      &   500    &      200  & 2\\
			Banana   &  5300       &    2     &      50  & 2 \\
			\hline
		\end{tabular}
	}
\vspace{0.6cm}
\end{table*}
\subsection{Hyperparameters Selection}
In this section, we will outline the parameters that we have selected for the proposed method. Instead of computing the full graph, we construct an $N_n$-nearest neighbor graph. To compute the graph weights, we use the Gaussian kernel function (\ref{gaussian}). The $\sigma$ parameter in the Gaussian kernel is a scalar, and it should be optimized so that the weight values are distributed among $[0,1]$. Having most weights very close to zero or very close to one is unfavorable. After the graph construction, we compute the family of persistent graph Laplacians as derived in Section \ref{persistent}. For each persistent Laplacian in the family, we compute its first $N_e$ eigenvalues and eigenvectors, where $N_e << N$. Overall, both $N_n$ and $N_e$ are hyperparameters that need tuning. Some other hyperparameters that might require tuning are the number of persistent Laplacians ($L_n$), the time step for solving the heat equation ($dt$), the constraint constant on the fidelity term ($\mu$), the maximum number of iterations ($N_t$), and the factor $C$ in the diffusion operator.

In this paper, we found the exact parameters used for the experiments using random search and outlined them in the Supplementary information.

\subsection{Performance and Discussion}

\begin{figure}
    \includegraphics[width = \textwidth ]{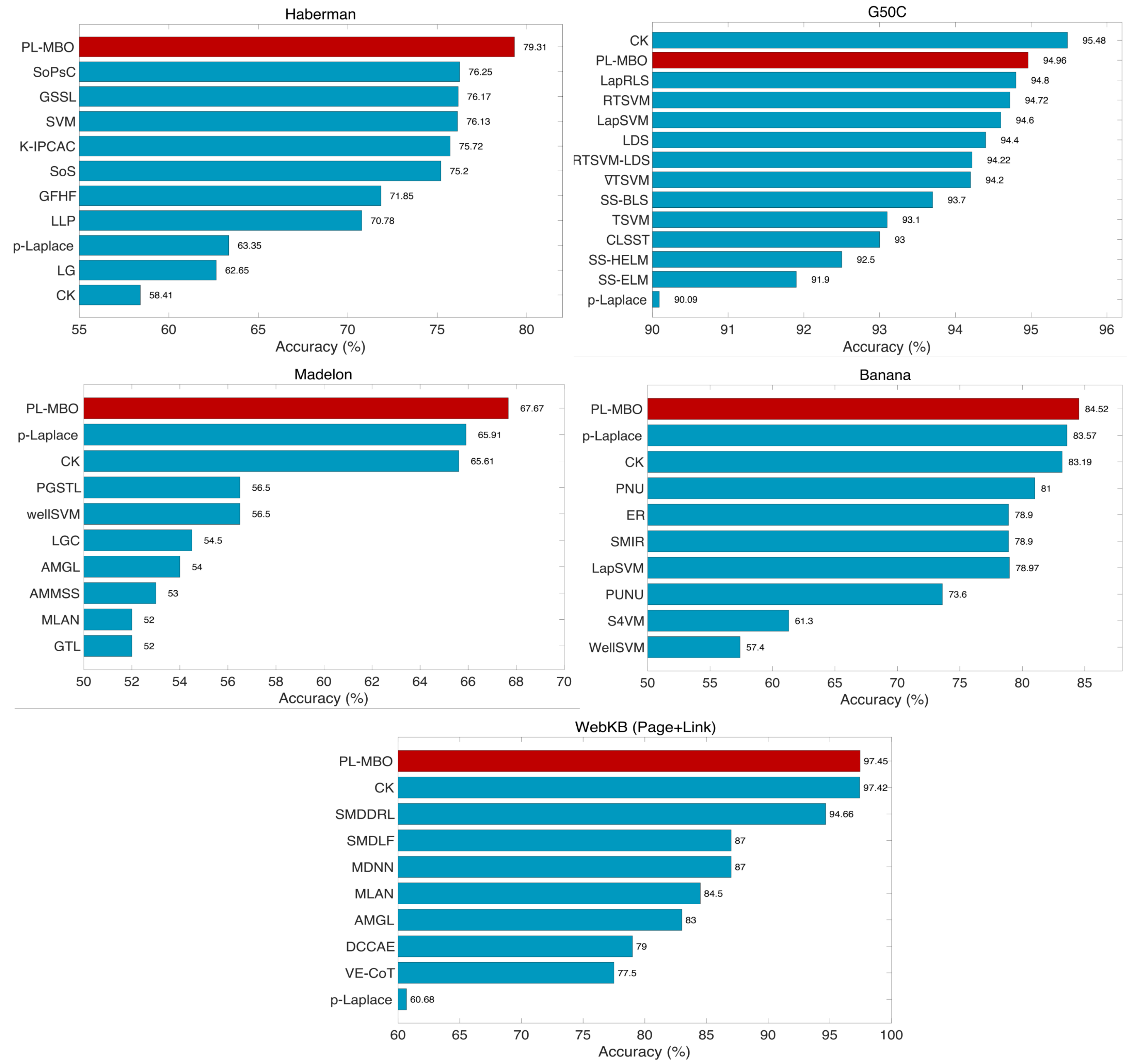}
    \caption{Accuracy comparison of the proposed PL-MBO algorithm with other methods on the Haberman, G50C, Madelon, Banana, and WebKB (Page+Link) data sets. The proposed method is indicated in red and the other methods are indicated in blue.}
\label{figcomp}
\end{figure}

The data sets that we used for our computational experiments are the G50C, WebKB, Madelon, Haberman, and Banana data sets. For all data sets, we consider classification accuracy as the main evaluation metric. The results of the experiments are shown in Figure \ref{figcomp}, with the proposed method outlined in red and the comparison methods outlined in blue. For all data sets except the G50C data set, our proposed method obtains higher accuracies compared to the comparison methods. For the G50C data set, our proposed method performs slightly lower than the Centered Kernel method (CK) \cite{ck}, but it performs better than the other comparison algorithms.

Overall, our proposed method performs well across a variety of datasets. For instance, while the G50C and Haberman datasets are relatively small compared to the other three, our method performs very well on both. Conversely, for larger datasets such as the Banana dataset, our proposed method has achieved superior performance compared to state-of-the-art comparison methods. Additionally, the WebKB and Madelon datasets have high sample dimensions, yet our proposed method still outperforms other methods on these high-dimensional datasets.

\subsection{Comparison Algorithms}

In this section, we compare our method with both recent and traditional methods. The results of the experiments and the comparison are shown in Figure \ref{figcomp}.

For all data sets, we compare our method, PL-MBO, to recent algorithms such as the Centered Kernel method (CK) \cite{ck} and $p$-Laplace learning ($p$-Laplace) \cite{PL}. For both comparison methods, we use 200 nearest neighbors. Moreover, for each data set, we have additional comparison methods described below.


For the G50C dataset, we also compare our proposed method with algorithms such as semi-supervised hierarchical extreme learning machines (SS-HELM) \cite{SS-HEML}, semi-supervised extreme learning machines (SS-ELM) \cite{SS-ELM}, semi-supervised broad learning system (SS-BLS) \cite{SS-BLS}, clustering (CLSST) \cite{CLSST}, transductive support vector machines (TSVM) \cite{gTSVM}, transductive support vector machines with training by gradient descent ($\nabla$TSVM) \cite{gTSVM}, low-density separation (LDS) \cite{gTSVM}, Laplacian support vector machines (LapSVM) \cite{LapSVM}, robust and fast transductive support vector machine (RTSVM, RTSVM-LDS) \cite{SVM}, and Laplacian regularized least squares (LapRLS) \cite{LapSVM1}. The results for CLSST are from \cite{CLSST}, the results for SS-BLS, SS-ELM, and SS-HELM are from \cite{SS-BLS}, the results for RTSVM, RTSVM-LDS are from \cite{SVM}, the results for LapRLS and LapSVM are from \cite{LapSVM}, and the results for TSVM, $\nabla$TSVM, and LDS are from \cite{gTSVM}. All of the comparison methods and the proposed method use 50 labeled elements. During the experiments, we randomly selected ten different sets of labeled data and computed the average accuracy for the final result. 

For the Haberman dataset, we also compare our proposed method with algorithms such as semi-supervised learning using Gaussian fields and harmonic functions (GFHF) \cite{GFHF}, graph-based semi-supervised learning (GSSL) \cite{GSSL}, logistic label propagation (LLP) \cite{LLP}, the graph-based Fisher kernel method (SoPsC) \cite{SoPsC}, learning with local and global consistency (LG) \cite{LGC}, the novel Fisher discriminant classifier (K-IPCAC) \cite{K-IPCAC}, the sum over paths covariance kernel (SoS) \cite{SoS}, and the training algorithm for optimal margin classifier (SVM) \cite{H-SVM}. The results for all these methods are from \cite{SoPsC}. All of the comparison methods and the proposed method use approximately 60 labeled examples. During the experiments, we randomly select fifty different sets of labeled data and compute the average accuracy for the final result.

For the WebKB data set, we also compare the accuracy of our algorithm with semi-supervised multi-view deep discriminant representation learning (SMDDRL) \cite{SMDDRL}, a semi-supervised multimodal deep learning framework (SMDLF) \cite{SMDLF}, auto-weighted multiple graph learning (AMGL) \cite{AMGL}, vertical ensemble co-training (VE-CoT) \cite{VE-COT}, multi-view learning with adaptive neighbors (MLAN) \cite{MLAN1}, a deep canonically correlated autoencoder (DCCAE) \cite{DCCAE}, and a multi-view discriminative neural network (MDNN) \cite{MDNN}. The results for all comparison methods are from the work \cite{SMDDRL}. All comparison methods consider ten percent of the data elements as labeled. During the experiments, we randomly select twenty different sets of labeled data and compute the average accuracy for the final result. 

For the Madelon dataset, we also compare the accuracy of our proposed algorithm with that of seven other semi-supervised methods: the automatic multi-model semi-supervised method (AMMSS) \cite{AMMSS}, semi-supervised learning using Gaussian fields and harmonic functions (GTL) \cite{GFHF}, progressive graph-based subspace transductive learning (PGSTL) \cite{PGSTL}, auto-weighted multiple graph learning method (AMGL) \cite{AMGL}, multi-view learning with adaptive neighbors (MLAN) \cite{MLAN}, weakly labeled support vector machine (WellSVM) \cite{WellSVM}, and local and global consistency method (LGC) \cite{LGC}. The results for all comparison methods are from \cite{PGSTL}. All of the comparison methods and the proposed method use 200 labeled elements. During the experiments, to obtain a fair comparison, we randomly select ten different sets of labeled data and compute the average accuracy for the final result.

For the Banana data set, we also compare against the following semi-supervised classification methods: the safe semi-supervised support vector machine (S4VM) \cite{S4VM}, entropy regularization (ER) \cite{g50c}, the Laplacian support vector machine (LapSVM) \cite{LapSVM}, squared-loss mutual information regularization (SMIR) \cite{SMIR}, the weakly labeled support vector machine (WellSVM) \cite{WellSVM}, and data classification from unlabeled positive and negative data (PNU) and (PUNU) \cite{PNU}. Results for all the comparison methods are from \cite{PNU}. All of the comparison methods, and the proposed method, use 50 labeled elements. During the experiments, we randomly select fifty different sets of labeled data and compute the average accuracy for the final result.

\section{Concluding Remarks} \label{Conclusion}
We present a novel topological graph-based semi-supervised method called PL-MBO by integrating persistent spectral graphs with an adaptation and graph-based modification of the classical Merriman-Bence-Osher (MBO) technique. This method is an efficient procedure that performs well with low label rates and small amounts of labeled data, which is crucial since labeled data is often scarce in many applications. The proposed algorithm is also adaptable for both small and large datasets. Experimental results on various benchmark datasets indicate that the proposed PL-MBO method outperforms other recent methods. In future work, we plan to explore integrating techniques such as various types of autoencoders for feature extraction, and to perform experiments with different types of molecular data. Overall, the proposed PL-MBO scheme is a powerful approach for data science when there are few labeled data elements.
  
\section{Supporting Information}
We present the optimal hyperparameters of the proposed method in Online Resources: Supporting Information. \\

\textbf{Code Availability:} The source code is available on Github at \url{https://github.com/kmerkurev/Persistent-Laplacian-Method} \\

\textbf{Conflict of Interest}: The authors declare no conflict of interest. \\

\section*{Acknowledgments}
This work was supported in part by NSF grant  DMS-2052983 and NIH grant  R01AI164266. 

\bibliography{Sources}

\begin{thebibliography}{10}

\bibitem{banana}
{B}anana {D}ata {S}et.
\newblock
  \url{https://github.com/SaravananJaichandar/MachineLearning/tree/master/Standard%20Classification%20Dataset/banana}.

\bibitem{webkb}
{CMU} {W}orld {W}ide {K}nowledge {B}ase ({W}eb{KB}) project.
\newblock \url{http://www.cs.cmu.edu/~webkb/}.

\bibitem{g_50c}
{G50C} {D}ata {S}et.
\newblock \url{http://vikas.sindhwani.org/datasets/ssl/}.

\bibitem{haberman}
{H}aberman {D}ata {S}et.
\newblock \url{https://archive.ics.uci.edu/ml/datasets/haberman%27s+survival}.

\bibitem{madelon}
{M}adelon {D}ata {S}et.
\newblock \url{https://github.com/godsylla/UCI-Madelon-Dataset}.

\bibitem{GF}
Amr Ahmed, Nino Shervashidze, Shravan Narayanamurthy, Vanja Josifovski, and
  Alexander~J. Smola.
\newblock Distributed large-scale natural graph factorization.
\newblock In {\em Proceedings of the 22nd International Conference on World
  Wide Web}, pages 37--48, 2013.

\bibitem{R-C}
Christopher~R. Anderson.
\newblock A {R}ayleigh--{C}hebyshev procedure for finding the smallest
  eigenvalues and associated eigenvectors of large sparse {H}ermitian matrices.
\newblock {\em Journal of Computational Physics}, 229(19):7477--7487, 2010.

\bibitem{IR}
Mikhail Belkin, Irina Matveeva, and Partha Niyogi.
\newblock Regularization and semi-supervised learning on large graphs.
\newblock In {\em 17th Annual Conference on Learning Theory}, pages 624--638,
  2004.

\bibitem{TR}
Mikhail Belkin, Irina Matveeva, and Partha Niyogi.
\newblock Tikhonov regularization and semi-supervised learning on large graphs.
\newblock In {\em IEEE International Conference on Acoustics, Speech, and
  Signal Processing}, volume~3, pages iii--1000, 2004.

\bibitem{LE}
Mikhail Belkin and Partha Niyogi.
\newblock Laplacian eigenmaps and spectral techniques for embedding and
  clustering.
\newblock {\em Advances in Neural Information Processing Systems}, 14, 2001.

\bibitem{MR1}
Mikhail Belkin, Partha Niyogi, and Vikas Sindhwani.
\newblock Manifold regularization: A geometric framework for learning from
  labeled and unlabeled examples.
\newblock {\em Journal of Machine Learning Research}, 7(11), 2006.

\bibitem{NE1}
Serge Belongie, Charless Fowlkes, Fan Chung, and Jitendra Malik.
\newblock Spectral partitioning with indefinite kernels using the {N}ystr{\"o}m
  extension.
\newblock In {\em 7th European Conference on Computer Vision}, pages 531--542,
  2002.

\bibitem{Bertozzi1}
Andrea~L Bertozzi and Arjuna Flenner.
\newblock Diffuse interface models on graphs for classification of high
  dimensional data.
\newblock {\em Multiscale Modeling \& Simulation}, 10(3):1090--1118, 2012.

\bibitem{H-SVM}
Bernhard~E. Boser, Isabelle~M. Guyon, and Vladimir~N. Vapnik.
\newblock A training algorithm for optimal margin classifiers.
\newblock In {\em Proceedings of the Fifth Annual Workshop on Computational
  Learning Theory}, pages 144--152, 1992.

\bibitem{AMMSS}
Xiao Cai, Feiping Nie, Weidong Cai, and Heng Huang.
\newblock Heterogeneous image features integration via multi-modal
  semi-supervised learning model.
\newblock In {\em Proceedings of the IEEE International Conference on Computer
  Vision}, pages 1737--1744, 2013.

\bibitem{PL}
Jeff Calder, Brendan Cook, Matthew Thorpe, and Dejan Slepcev.
\newblock Poisson learning: Graph based semi-supervised learning at very low
  label rates.
\newblock In {\em International Conference on Machine Learning}, pages
  1306--1316. PMLR, 2020.

\bibitem{GSSL}
Gustavo Camps-Valls, Tatyana V.~Bandos Marsheva, and Dengyong Zhou.
\newblock Semi-supervised graph-based hyperspectral image classification.
\newblock {\em IEEE Transactions on Geoscience and Remote Sensing},
  45(10):3044--3054, 2007.

\bibitem{GraRep}
Shaosheng Cao, Wei Lu, and Qiongkai Xu.
\newblock Gra{R}ep: Learning graph representations with global structural
  information.
\newblock In {\em Proceedings of the 24th ACM International on Conference on
  Information and Knowledge Management}, pages 891--900, 2015.

\bibitem{SSL}
Olivier Chapelle, Bernhard Schölkopf, and Alexander Zien, editors.
\newblock {\em Semi-Supervised Learning}.
\newblock The MIT Press, 2006.

\bibitem{gTSVM}
Olivier Chapelle and Alexander Zien.
\newblock Semi-supervised classification by low density separation.
\newblock In {\em International Workshop on Artificial Intelligence and
  Statistics}, pages 57--64, 2005.

\bibitem{HARP}
Haochen Chen, Bryan Perozzi, Yifan Hu, and Steven Skiena.
\newblock {HARP}: Hierarchical representation learning for networks.
\newblock In {\em Proceedings of the AAAI Conference on Artificial
  Intelligence}, volume~32, 2018.

\bibitem{chen2022persistent}
Jiahui Chen, Yuchi Qiu, Rui Wang, and Guo-Wei Wei.
\newblock Persistent {L}aplacian projected {O}micron {BA}. 4 and {BA}. 5 to
  become new dominating variants.
\newblock {\em Computers in Biology and Medicine}, 151:106262, 2022.

\bibitem{PGSTL}
Long Chen and Zhi Zhong.
\newblock Progressive graph-based subspace transductive learning for
  semi-supervised classification.
\newblock {\em IET Image Processing}, 13(14):2753--2762, 2019.

\bibitem{chen}
Yunmei Chen and Xiaojing Ye.
\newblock Projection onto a simplex.
\newblock {\em arXiv preprint arXiv:1101.6081}, 2011.

\bibitem{SMDLF}
Yanhua Cheng, Xin Zhao, Rui Cai, Zhiwei Li, Kaiqi Huang, Yong Rui, et~al.
\newblock Semi-supervised multimodal deep learning for {RGB-D} object
  recognition.
\newblock In {\em International Joint Conference on Artificial Intelligence},
  pages 3345--3351, 2016.

\bibitem{SVM}
Corinna Cortes and Vladimir Vapnik.
\newblock Support-vector networks.
\newblock {\em Machine Learning}, 20:273--297, 1995.

\bibitem{edelsbrunner2008persistent}
Herbert Edelsbrunner and John Harer.
\newblock Persistent homology-a survey.
\newblock {\em Contemporary Mathematics}, 453(26):257--282, 2008.

\bibitem{NE2}
Charless Fowlkes, Serge Belongie, and Jitendra Malik.
\newblock Efficient spatiotemporal grouping using the {N}ystr{\"o}m method.
\newblock In {\em Proceedings of the 2001 IEEE Computer Society Conference on
  Computer Vision and Pattern Recognition}, volume~1, pages I--I, 2001.

\bibitem{G-C}
Cristina Garcia-Cardona, Ekaterina Merkurjev, Andrea~L. Bertozzi, Arjuna
  Flenner, and Allon~G. Percus.
\newblock Multiclass data segmentation using diffuse interface methods on
  graphs.
\newblock {\em IEEE Transactions on Pattern Analysis and Machine Intelligence},
  36(8):1600--1613, 2014.

\bibitem{DGL}
Chen Gong, Tongliang Liu, Dacheng Tao, Keren Fu, Enmei Tu, and Jie Yang.
\newblock Deformed graph {L}aplacian for semi-supervised learning.
\newblock {\em IEEE Transactions on Neural Networks and Learning Systems},
  26(10):2261--2274, 2015.

\bibitem{g50c}
Yves Grandvalet and Yoshua Bengio.
\newblock Semi-supervised learning by entropy minimization.
\newblock {\em Advances in Neural Information Processing Systems}, 17, 2004.

\bibitem{NV}
Aditya Grover and Jure Leskovec.
\newblock node2vec: {S}calable feature learning for networks.
\newblock In {\em Proceedings of the 22nd ACM SIGKDD International Conference
  on Knowledge Discovery and Data Mining}, pages 855--864, 2016.

\bibitem{book}
Allen Hatcher.
\newblock {\em Algebraic topology}.
\newblock 2005.

\bibitem{Nicole}
Nicole Hayes, Ekaterina Merkurjev, and Guo-Wei Wei.
\newblock Integrating transformer and autoencoder techniques with spectral
  graph algorithms for the prediction of scarcely labeled molecular data.
\newblock {\em Computers in Biology and Medicine}, 153:106479, 2023.

\bibitem{SS-ELM}
Gao Huang, Shiji Song, Jatinder~ND Gupta, and Cheng Wu.
\newblock Semi-supervised and unsupervised extreme learning machines.
\newblock {\em IEEE Transactions on Cybernetics}, 44(12):2405--2417, 2014.

\bibitem{auction}
Matt Jacobs, Ekaterina Merkurjev, and Selim Esedoḡlu.
\newblock Auction dynamics: A volume constrained mbo scheme.
\newblock {\em Journal of Computational Physics}, 354:288--310, 2018.

\bibitem{SMDDRL}
Xiaodong Jia, Xiao-Yuan Jing, Xiaoke Zhu, Songcan Chen, Bo~Du, Ziyun Cai,
  Zhenyu He, and Dong Yue.
\newblock Semi-supervised multi-view deep discriminant representation learning.
\newblock {\em IEEE Transactions on Pattern Analysis and Machine Intelligence},
  43(7):2496--2509, 2020.

\bibitem{VE-COT}
Gilad Katz, Cornelia Caragea, and Asaf Shabtai.
\newblock Vertical ensemble co-training for text classification.
\newblock {\em ACM Transactions on Intelligent Systems and Technology (TIST)},
  9(2):1--23, 2017.

\bibitem{GCN}
Thomas~N Kipf and Max Welling.
\newblock Semi-supervised classification with graph convolutional networks.
\newblock {\em Proceedings of 5th International Conference on Learning
  Representations}, pages 1--14, 2017.

\bibitem{LLP}
Takumi Kobayashi, Kenji Watanabe, and Nobuyuki Otsu.
\newblock Logistic label propagation.
\newblock {\em Pattern Recognition Letters}, 33(5):580--588, 2012.

\bibitem{LI}
Ruoyu Li, Sheng Wang, Feiyun Zhu, and Junzhou Huang.
\newblock Adaptive graph convolutional neural networks.
\newblock In {\em Proceedings of the AAAI Conference on Artificial
  Intelligence}, volume~32, 2018.

\bibitem{WellSVM}
Yu-Feng Li, Ivor~W Tsang, James~T Kwok, and Zhi-Hua Zhou.
\newblock Convex and scalable weakly labeled svms.
\newblock {\em Journal of Machine Learning Research}, 14(7), 2013.

\bibitem{S4VM}
Yu-Feng Li and Zhi-Hua Zhou.
\newblock Towards making unlabeled data never hurt.
\newblock {\em IEEE Transactions on Pattern Analysis and Machine Intelligence},
  37(1):175--188, 2014.

\bibitem{AGR}
Wei Liu, Junfeng He, and Shih-Fu Chang.
\newblock Large graph construction for scalable semi-supervised learning.
\newblock In {\em Proceedings of the 27th International Conference on Machine
  Learning}, pages 679--686, 2010.

\bibitem{ck}
Xiaoyi Mai and Romain Couillet.
\newblock Random matrix-inspired improved semi-supervised learning on graphs.
\newblock In {\em International Conference on Machine Learning}, 2018.

\bibitem{SoS}
Amin Mantrach, Luh Yen, Jerome Callut, Kevin Francoisse, Masashi Shimbo, and
  Marco Saerens.
\newblock The sum-over-paths covariance kernel: A novel covariance measure
  between nodes of a directed graph.
\newblock {\em IEEE Transactions on Pattern Analysis and Machine Intelligence},
  32(6):1112--1126, 2009.

\bibitem{LapSVM}
Stefano Melacci and Mikhail Belkin.
\newblock Laplacian support vector machines trained in the primal.
\newblock {\em Journal of Machine Learning Research}, 12(3), 2011.

\bibitem{PLT}
Facundo M{\'e}moli, Zhengchao Wan, and Yusu Wang.
\newblock Persistent {L}aplacians: Properties, algorithms and implications.
\newblock {\em SIAM Journal on Mathematics of Data Science}, 4(2):858--884,
  2022.

\bibitem{meng}
Zhaoyi Meng, Ekaterina Merkurjev, Alice Koniges, and Andrea~L Bertozzi.
\newblock Hyperspectral image classification using graph clustering methods.
\newblock {\em Image Processing On Line}, 7:218--245, 2017.

\bibitem{PLB}
Zhenyu Meng and Kelin Xia.
\newblock Persistent spectral-based machine learning ({P}er{S}pect {ML}) for
  protein-ligand binding affinity prediction.
\newblock {\em Science Advances}, 7(19):eabc5329, 2021.

\bibitem{clouds}
Ekaterina Merkurjev.
\newblock A fast graph-based data classification method with applications to
  {3D} sensory data in the form of point clouds.
\newblock {\em Pattern Recognition Letters}, 136:154--160, 2020.

\bibitem{pagerank}
Ekaterina Merkurjev, Andrea~L Bertozzi, and Fan Chung.
\newblock A semi-supervised heat kernel pagerank mbo algorithm for data
  classification.
\newblock {\em Communications in Mathematical Sciences}, 16(5):1241--1265,
  2018.

\bibitem{Merkurjev1}
Ekaterina Merkurjev, Cristina Garcia-Cardona, Andrea~L Bertozzi, Arjuna
  Flenner, and Allon~G Percus.
\newblock Diffuse interface methods for multiclass segmentation of
  high-dimensional data.
\newblock {\em Applied Mathematics Letters}, 33:29--34, 2014.

\bibitem{Merkurjev2}
Ekaterina Merkurjev, Tijana Kostic, and Andrea~L Bertozzi.
\newblock An {MBO} scheme on graphs for classification and image processing.
\newblock {\em SIAM Journal on Imaging Sciences}, 6(4):1903--1930, 2013.

\bibitem{MLL}
Ekaterina Merkurjev, Duc~Duy Nguyen, and Guo-Wei Wei.
\newblock Multiscale {L}aplacian {L}earning.
\newblock {\em Applied Intelligence}, pages 1--20, 2022.

\bibitem{sunu}
Ekaterina Merkurjev, Justin Sunu, and Andrea~L Bertozzi.
\newblock Graph {MBO} method for multiclass segmentation of hyperspectral
  stand-off detection video.
\newblock In {\em IEEE International Conference on Image Processing}, pages
  689--693, 2014.

\bibitem{originalMBO}
Barry Merriman, James~K. Bence, and Stanley~J. Osher.
\newblock Motion of multiple junctions: A level set approach.
\newblock {\em Journal of Computational Physics}, 112(2):334--363, 1994.

\bibitem{MLAN}
Feiping Nie, Guohao Cai, Jing Li, and Xuelong Li.
\newblock Auto-weighted multi-view learning for image clustering and
  semi-supervised classification.
\newblock {\em IEEE Transactions on Image Processing}, 27(3):1501--1511, 2017.

\bibitem{MLAN1}
Feiping Nie, Guohao Cai, and Xuelong Li.
\newblock Multi-view clustering and semi-supervised classification with
  adaptive neighbours.
\newblock In {\em Proceedings of the AAAI Conference on Artificial
  Intelligence}, volume~31, 2017.

\bibitem{AMGL}
Feiping Nie, Jing Li, and Xuelong Li.
\newblock Parameter-free auto-weighted multiple graph learning: a framework for
  multiview clustering and semi-supervised classification.
\newblock In {\em International Joint Conference on Artificial Intelligence},
  pages 1881--1887, 2016.

\bibitem{SMIR}
Gang Niu, Wittawat Jitkrittum, Bo~Dai, Hirotaka Hachiya, and Masashi Sugiyama.
\newblock Squared-loss mutual information regularization: A novel
  information-theoretic approach to semi-supervised learning.
\newblock In {\em International Conference on Machine Learning}, pages 10--18,
  2013.

\bibitem{MDNN}
Vahid Noroozi, Sara Bahaadini, Lei Zheng, Sihong Xie, Weixiang Shao, and
  Phillip~S. Yu.
\newblock Semi-supervised deep representation learning for multi-view problems.
\newblock In {\em IEEE International Conference on Big Data}, pages 56--64,
  2018.

\bibitem{HOPE}
Mingdong Ou, Peng Cui, Jian Pei, Ziwei Zhang, and Wenwu Zhu.
\newblock Asymmetric transitivity preserving graph embedding.
\newblock In {\em ACM SIGKDD International Conference on Knowledge Discovery
  and Data Mining}, pages 1105--1114, 2016.

\bibitem{DeepWalk}
Bryan Perozzi, Rami Al-Rfou, and Steven Skiena.
\newblock Deepwalk: Online learning of social representations.
\newblock In {\em ACM SIGKDD International Conference on Knowledge Discovery
  and Data Mining}, pages 701--710, 2014.

\bibitem{LLE}
Sam~T Roweis and Lawrence~K. Saul.
\newblock Nonlinear dimensionality reduction by locally linear embedding.
\newblock {\em Science}, 290(5500):2323--2326, 2000.

\bibitem{K-IPCAC}
Alessandro Rozza, Gabriele Lombardi, Elena Casiraghi, and Paola Campadelli.
\newblock Novel {F}isher discriminant classifiers.
\newblock {\em Pattern Recognition}, 45(10):3725--3737, 2012.

\bibitem{SoPsC}
Alessandro Rozza, Mario Manzo, and Alfredo Petrosino.
\newblock A novel graph-based {F}isher kernel method for semi-supervised
  learning.
\newblock In {\em International Conference on Pattern Recognition}, pages
  3786--3791, 2014.

\bibitem{PNU}
Tomoya Sakai, Marthinus~Christoffel Plessis, Gang Niu, and Masashi Sugiyama.
\newblock Semi-supervised classification based on classification from positive
  and unlabeled data.
\newblock In {\em International Conference on Machine Learning}, pages
  2998--3006, 2017.

\bibitem{CLSST}
Emanuele Sansone, Andrea Passerini, and Francesco De~Natale.
\newblock Classtering: Joint classification and clustering with mixture of
  factor analysers.
\newblock In {\em Proceedings of the European Conference on Artificial
  Intelligence}, pages 1089--1095, 2016.

\bibitem{SS-HEML}
Qingshan She, Bo~Hu, Zhizeng Luo, Thinh Nguyen, and Yingchun Zhang.
\newblock A hierarchical semi-supervised extreme learning machine method for
  {EEG} recognition.
\newblock {\em Medical \& Biological Engineering \& Computing}, 57:147--157,
  2019.

\bibitem{LapSVM1}
Vikas Sindhwani, Partha Niyogi, and Mikhail Belkin.
\newblock Beyond the point cloud: from transductive to semi-supervised
  learning.
\newblock In {\em Proceedings of the International Conference on Machine
  Learning}, pages 824--831, 2005.

\bibitem{LINE}
Jian Tang, Meng Qu, Mingzhe Wang, Ming Zhang, Jun Yan, and Qiaozhu Mei.
\newblock {LINE}: Large-scale information network embedding.
\newblock In {\em Proceedings of the International Conference on World Wide
  Web}, pages 1067--1077, 2015.

\bibitem{kd-tree}
A.~Vedaldi and B.~Fulkerson.
\newblock {VLFeat}: An open and portable library of computer vision algorithms.
\newblock \url{http://www.vlfeat.org/}, 2008.

\bibitem{GAN1}
Petar Velickovic, Guillem Cucurull, Arantxa Casanova, Adriana Romero, Pietro
  Lio, and Yoshua Bengio.
\newblock Graph attention networks.
\newblock {\em Proceedings of the International Conference on Learning
  Representations}, 1050(20):1--12, 2018.

\bibitem{wang}
Rui Wang, Duc~Duy Nguyen, and Guo-Wei Wei.
\newblock Persistent spectral graph.
\newblock {\em International Journal for Numerical Methods in Biomedical
  Engineering}, 36(9):e3376, 2020.

\bibitem{DCCAE}
Weiran Wang, Raman Arora, Karen Livescu, and Jeff Bilmes.
\newblock On deep multi-view representation learning.
\newblock In {\em International Conference on Machine Learning}, pages
  1083--1092, 2015.

\bibitem{GAN3}
Xiao Wang, Houye Ji, Chuan Shi, Bai Wang, Yanfang Ye, Peng Cui, and Philip~S.
  Yu.
\newblock Heterogeneous graph attention network.
\newblock In {\em The World Wide Web Conference}, pages 2022--2032, 2019.

\bibitem{MR2}
Zenglin Xu, Irwin King, Michael Rung-Tsong Lyu, and Rong Jin.
\newblock Discriminative semi-supervised feature selection via manifold
  regularization.
\newblock {\em IEEE Transactions on Neural Networks}, 21(7):1033--1047, 2010.

\bibitem{Planetoid}
Zhilin Yang, William Cohen, and Ruslan Salakhudinov.
\newblock Revisiting semi-supervised learning with graph embeddings.
\newblock In {\em International Conference on Machine Learning}, pages 40--48,
  2016.

\bibitem{GAN2}
Jiani Zhang, Xingjian Shi, Junyuan Xie, Hao Ma, Irwin King, and Dit-Yan Yeung.
\newblock Ga{AN}: Gated attention networks for learning on large and
  spatiotemporal graphs.
\newblock {\em Proceedings of the Thirty-Fourth Conference on Uncertainty in
  Artificial Intelligence}, pages 339--349, 2018.

\bibitem{SS-BLS}
Huimin Zhao, Jianjie Zheng, Wu~Deng, and Yingjie Song.
\newblock Semi-supervised broad learning system based on manifold
  regularization and broad network.
\newblock {\em IEEE Transactions on Circuits and Systems I: Regular Papers},
  67(3):983--994, 2020.

\bibitem{LGC}
Dengyong Zhou, Olivier Bousquet, Thomas Lal, Jason Weston, and Bernhard
  Sch{\"o}lkopf.
\newblock Learning with local and global consistency.
\newblock {\em Advances in Neural Information Processing Systems}, 16, 2003.

\bibitem{DR}
Dengyong Zhou, Jiayuan Huang, and Bernhard Sch{\"o}lkopf.
\newblock Learning from labeled and unlabeled data on a directed graph.
\newblock In {\em Proceedings of International Conference on Machine Learning},
  pages 1036--1043, 2005.

\bibitem{LP}
Xiaojin Zhu and Zoubin Ghahramani.
\newblock Learning from labeled and unlabeled data with label propagation.
\newblock {\em CMU CALD Tech Report CMU-CALD-02-107}, 2002.

\bibitem{GFHF}
Xiaojin Zhu, Zoubin Ghahramani, and John Lafferty.
\newblock Semi-supervised learning using {G}aussian fields and harmonic
  functions.
\newblock In {\em International Conference on Machine Learning}, volume~3, page
  912, 2003.

\bibitem{zomorodian2004computing}
Afra Zomorodian and Gunnar Carlsson.
\newblock Computing persistent homology.
\newblock In {\em Proceedings of the Twentieth Annual Symposium on
  Computational Geometry}, pages 347--356, 2004.

\end{thebibliography}
\bibliographystyle{plain}

\end{document}